# Accelerating hydrodynamic simulations of urban drainage systems with physics-guided machine learning

Rocco Palmitessa[1], Morten Grum[2], Allan Peter Engsig-Karup[3], Roland Löwe[1,*]

[1]Technical University of Denmark, Department of Environmental and Resources Engineering, Section of Climate and Monitoring, Miljøvej B115, 2800 Kgs. Lyngby, Denmark; rocp@env.dtu.dk, rolo@env.dtu.dk
[2]WaterZerv, Fjenneslevvej 23, st., 2700 Brønshøj, Denmark; mortengrum@waterzerv.com
[3]Technical University of Denmark, Department of Applied Mathematics and Computer Science, Section of Scientific Computing, Asmussens Allé, 303B, 2800 Kgs. Lyngby, Denmark; apek@dtu.dk
*corresponding author

### Abstract

We propose and demonstrate a new approach for fast and accurate surrogate modelling of urban drainage system hydraulics based on physics-guided machine learning. The surrogates are trained against a limited set of simulation results from a hydrodynamic (HiFi) model. Our approach reduces simulation times by one to two orders of magnitude compared to a HiFi model. It is thus slower than e.g. conceptual hydrological models, but it enables simulations of water levels, flows and surcharges in all nodes and links of a drainage network and thus largely preserves the level of detail provided by HiFi models. Comparing time series simulated by the surrogate and the HiFi model, $R^2$ values in the order of 0.9 are achieved. Surrogate training times are currently in the order of one hour. However, they can likely be reduced through the application of transfer learning and graph neural networks. Our surrogate approach will be useful for interactive workshops in initial design phases of urban drainage systems, as well as for real time applications. In addition, our model formulation is generic and future research should investigate its application for simulating other water systems.

# 1   Introduction

Computational modelling is essential in all phases of the operation of urban drainage systems (UDS), from design to monitoring and control. Physics-based deterministic models are widely used around the world. These models solve Saint Venant's system of equations in small time steps for each pipe element (Abbott and Ionescu, 1967). We will refer to this approach as "high-fidelity models". High-fidelity (Hifi) models are directly linked to physical system characteristics such as pipe diameter and length, and both water levels and flows are simulated in a realistic manner which makes them attractive to practitioners. However, high simulation times limit stakeholder involvement in the design phase (Leskens et al., 2014), as well as the exploration of impacts of deep uncertainties about future climates and urban developments (Löwe et al., 2020). In addition, real-time applications such as warning and control usually require shorter simulation times than what is feasible with HiFi models.

To circumvent these problems, hydrologists have developed a number of so-called lower-fidelity surrogates (Razavi et al., 2012). In particular, a variety of conceptual (or lumped) hydrological models were developed for UDS (Burger et al., 2016; Kroll et al., 2017; Machac et al., 2016; Moreno-Rodenas et al., 2018; Thrysøe et al., 2019). All these approaches reduce computation times by orders of magnitude. However, they achieve speedup at the cost of simplifying the simulated processes or lowering the spatiotemporal resolution. Typically, only flows are simulated (not water levels), and only few selected locations in the pipe network are considered. The practical consequences are that the existing surrogates are valid only for a limited range of purposes (e.g. sewer overflow but not flood risk), are not straightforward to set up automatically from pipe databases, and extrapolate poorly to a changing drainage system structure. This limitation is not present for cellular automata approaches that only simplify the mathematical description of the processes but preserve the level of detail of the model (Austin et al., 2014). However, similarly small simulation time steps as with a numerical simulator must be selected. Limited speed-ups in the order of factor 5 where therefore achieved.

Machine learning approaches have gained traction in hydrology. They are frequently applied in an input-output setting, including settings where physical system characteristics are used as inputs. The resulting models can be transferable between catchments as demonstrated, for example, for rainfall runoff modelling (Kratzert et al., 2019) and flood predictions (Bentivoglio et al., 2022; Löwe et al., 2021). Efforts have also been made to ensure conservation of mass (Hoedt et al., 2021). Both in rainfall-runoff modelling and in hydraulics, data-driven approaches have also been integrated into numerical solvers, with the aim of either simplifying numerical process equations, uncovering unknown relationships or achieving better model fits. Techniques such as genetic programming (Babovic and Abbott, 1997; Danandeh Mehr et al., 2018), neural networks (Höge et al., 2022) and curve fitting (Jamali et al., 2021) were used. While this approach can yield better fit with data and preserves physical interpretability, it is subject to similar limitations as the cellular automata approach in terms of computational performance. For example, (Jamali et al., 2021) achieved very limited reductions in simulation time for a problem involving solutions of the 2D shallow water equations.

In summary, there is a gap for surrogate approaches that simulate the full hydraulics for the often many thousand states included in a HiFi model, while still achieving substantial reductions in simulation time. In this paper, we argue that scientific (or physics-informed) machine learning (Karniadakis et al., 2021; Willard et al., 2020) is an avenue worth investigating and we develop a proof of concept for this purpose. In contrast to the surrogate approaches presented above, scientific machine learning aims to develop data-driven simulation models for any dynamical system, which has been demonstrated for a range of differential and partial differential equation systems (Geneva and Zabaras, 2020; Raissi et al., 2019). It exploits the computational efficiency of neural networks to create models that include large numbers of state variables. This ability is relevant for UDS, where we often want to simulate water levels and flows in hundreds or thousands of links and nodes. Further, scientific machine learning models learn interrelations between multiple states (e.g. water levels in neighboring nodes), and enable the exploitation of time series trends to increase simulation accuracy (Geneva and Zabaras, 2020; Ren et al., 2022). For hydraulics, these properties may enable the consideration of larger simulation time steps than presented in the literature so far. Finally, as outlined in the review papers by (Karniadakis et al., 2021; Willard et al., 2020), scientific machine learning enables multiple ways to incorporate physical system understanding into a data-driven model, e.g. in the form of physics-guided model architectures, mathematical constraints that enforce physical behavior, or penalizing the model during training if simulations are not aligned with known differential equations governing a system (Wang et al., 2021).

For this paper, our ambition is to:

1. Develop a proof of concept for simulating the hydraulics of UDS in high detail using scientific machine learning. To our knowledge, this has not been attempted before. Specifically, we build on generalized residue networks (Chen and Xiu, 2021). As suggested by (Garzón et al., 2022), we incorporate inductive bias by designing a model architecture that is aligned with physical intuition about the hydraulic behavior of the pipe system.
2. Illustrate potentials as well as current bottlenecks and pitfalls when applying these techniques in a realistic setting with time-varying rainfall inputs and complex pipe flow dynamics
3. Draw up further research directions based on current developments in the literature.

## 2 Material and methods

### 2.1 Principal concepts

Our aim is to construct a fast surrogate model for 1D hydrodynamics in pipes that enables the simulation of both water levels and flows in all nodes and links of an UDS, much like existing HiFi models. HiFi models for UDS's typically simulate catchment runoff independently. Subsequently, it is routed through the nodes into the drainage network where hydraulics are simulated by numerically solving the Saint Venant equations. The runoff simulation typically has marginal computational cost, while the network simulation is computationally expensive. Thus, we propose a surrogate model of the drainage network hydraulics only. The hydraulic states of interest for each node in the drainage network are the water level in the node ($h$), the flow in upstream and downstream links ($Q$), as well as excess flows ($Q_w$), for which we later will distinguish between frequently activated overflows ($Q_{w,Overflow}$), and infrequently activated surcharges that cause pluvial flooding ($Q_{w,Surcharge}$) (Figure 1a).

Because we focus only on the network component, the surrogate model is given the runoff $R$ as input and only predicts $h, Q$ and $Q_w$ (Figure 1b). Moreover, the model is structured in an autoregressive fashion: The model predicts how the hydraulic states change from one time step to another, given a vector of runoff inputs during this time step. The predicted states are then used as the starting point for the next time step. Thus, the model learns how the hydraulic states change over a time interval $\Delta t$, considering surface runoff as a boundary condition. Once initialized, the surrogate can independently simulate time periods of any duration, given a series of runoff inputs.

In this paper, we will consider a surrogate time step $\Delta t$ of 1 minute, which is well above the stability criteria for the HiFi model. The surrogate is trained against a limited set of simulation results from a HiFi model (so-called labeled data).

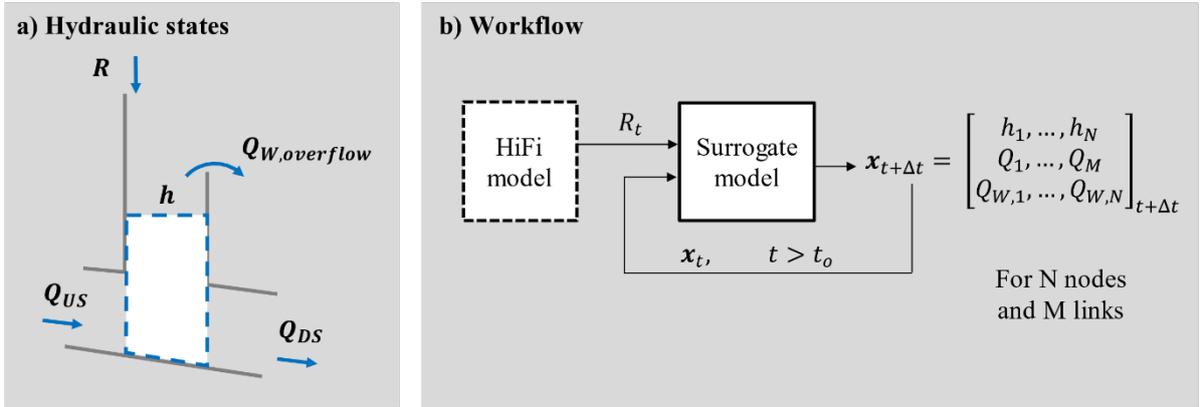

Figure 1. a): hydraulic states of interest in the urban drainage model are the catchments runoff ($R$), the water level in nodes ($h$), the flow ($Q$) in the links upstream and downstream to the node, and the excess flow ($Q_w$). b): in the proposed workflow, the runoff and the initial states simulated with the HiFi model are given as input to the surrogate model; after initialization, the surrogate model predictions of water levels and flows at the previous time step are fed back as input for the length of the output window.

## 2.2 Model formulation

We start by introducing the concept of generalized residue networks, then we provide a model formulation for urban drainage systems, and finally we introduce physical system understanding into this model.

### 2.2.1 Generalized residue networks (gResNet)

Consider a dynamical system posed as an initial value problem with state vector **x** that changes over time t:

$$\frac{d\boldsymbol{x}}{dt} = f(\boldsymbol{x}), \; t > t_o, \; \boldsymbol{x}(t_o) = \boldsymbol{x}_o. \tag{1}$$

We define the flow map as a real-valued function that maps how the state vector of the system changes from time $t$ to $t + \Delta t$. (Qin et al., 2019) suggested that this flow map can be approximated by the function

$$\boldsymbol{x}_{t+\Delta t} = \boldsymbol{x}_t + N(\boldsymbol{x}_t, \boldsymbol{\Theta}), \tag{2}$$

where $N$ is a deep feed forward neural network with parameters $\boldsymbol{\Theta}$. The principal idea of this approach is that the solution of the differential equation system in Eq. 1 is approximated in discrete time steps

$\Delta t$ using an identity mapping of the model states and a deep neural network that learns the "residue" from data, thereby capturing the system dynamics without knowing the true $f(x)$ a priori. Initializing from some starting states $x_o$, Eq. 2 can be used to simulate the system forward in time. (Chen and Xiu, 2021) extended the approach by replacing the identity mapping in Eq. 2 with a flexible operator mapping $x_{t+\Delta t} \approx \mathcal{L}(x_t)$. $\mathcal{L}$ can in principle be any model approximation for the system, but with inspiration from dynamic mode decomposition techniques (Schmid, 2010) they suggested that a feed forward neural network $L$ with a single hidden layer is a strong candidate for approximating the state changes from one time step to another. This leads to the formulation of generalized residue networks (gResNet):

$$x_{t+\Delta t} = L(x_t, \Theta_L) + N(x_t, \Theta_N). \tag{3}$$

### 2.2.2 Generalized residue networks for UDS

The model formulation in Eq. 3 is attractive for the simulation of urban drainage networks because it allows for the consideration of a large number of states in the vector $x$, because we can consider various formulations of the neural network $N$ that reflect different system complexities, and because we can impose physical constraints on model predictions that will be introduced in the next section. Assuming that we want to use Eq. 3 to simulate an UDS with $N$ nodes and $M$ links, we consider a state vector that includes the water levels $h$ in all nodes, the flows $Q$ through all links, and the excess flows $Q_w$ from all nodes (Figure 1a):

$$x_t = [h_1, \dots, h_N, Q_1, \dots, Q_M, Q_{w,1}, \dots, Q_{w,N}]. \tag{4}$$

The state changes from one time step to another depend on the boundary conditions imposed on the system, i.e. on the vector of runoff volumes $R_t = [R_1, \dots, R_N]$ that enter the nodes in the considered time interval. We therefore consider $R_t$ as an additional input to the deep neural network $N$, while we preserve the purely state dependent formulation of the prior model $L$:

$$x_{t+\Delta t} = L(x_t, \Theta_L) + N(x_t, R_t, \Theta_N). \tag{5}$$

Starting from some initial state values $x_0$, this model simulates water levels, pipe flows and surcharge flows in time intervals $\Delta t$ for each node and each link for which the model is trained.

### 2.2.3 Physics-constrained generalized residue networks for UDS

Predictions from the model in Eq. 5 are in no way constrained to a range that is physically reasonable. (Beucler et al., 2021) suggested in a similar setting that one way of ensuring physically appropriate behavior of the model is to introduce so-called constraint layers $C$. These can be interpreted as a post-processing step after each prediction, and calculate certain model states based on a physical understanding of the system:

$$x_{t+\Delta t} = C\big(L(x_t, \Theta_L) + N(x_t, R_t, \Theta_N)\big). \tag{6}$$

An import learning from initial experiments with Eq. 5 was that surcharge flows $Q_{w,Surcharge}$ were difficult to learn for a purely data-driven model. However, surcharge flows occur only when the water level in a node reaches the ground level, and the connected pipes exceed their capacity. In this situation, the water volume stored in the pipe network remains constant, and $Q_{w,Surcharge}$ can be computed from a mass balance calculation that is performed locally for each node. Therefore, we suggest a revised model formulation where the state vector (Eq. 4) is reduced to only include water levels and pipe flows:

$$\boldsymbol{x}_t = [h_1, \ldots, h_N, Q_1, \ldots, Q_M]. \tag{7}$$

Surcharges are subsequently calculated in a "constraint layer" as the difference between all the inflows to a node $i$ from upstream pipe elements (considering the link flow values predicted by the gResNet) and runoff, and the outflow to all downstream pipe elements. In addition, we enforce a minimum of 0 for surcharge flows, because we perform HiFi simulations in "spilling configuration", i.e. surcharging water does not reenter the drainage system. This configuration was shown to yield more realistic representations of pressure levels in intense rain storms than the widely used "ponding configuration" (Jamali et al., 2018):

$$Q_{w,i} = \max\left(\sum_j Q_{US,i,j} - \sum_k Q_{DS,i,k} + R_i, 0\right). \tag{8}$$

Note that while we have removed $Q_w$ from the state vector, the loss function for training the parameter sets $\boldsymbol{\Theta}_L$ and $\boldsymbol{\Theta}_N$ is still computed on all the states $h$, $Q$ and $Q_w$ (Section 2.3). Thus, the mass balance computation affects the parameter estimates of the gResNet.

## 2.3 Surrogate training

The proposed model includes parameter sets $\boldsymbol{\Theta}_L$ and $\boldsymbol{\Theta}_N$ that need to be estimated. For this purpose, we simulate rainfall series of limited duration in a HiFi model and extract time series of the simulated hydraulic state variables for all nodes and links in 1 minute resolution. The surrogate parameters are estimated by minimizing the mean squared error (MSE) between the "labels" generated by the HiFi model and the corresponding hydraulic states simulated by the surrogate. All states are scaled to the range [0,1] before computing MSE, to ensure that all state variables have approximately equal impact on the parameter estimates. Scaling is performed using a min-max approach (Pedregosa et al., 2011), and the scaling range is defined individually for each state variable, based on the minimum and maximum values observed in the HiFi simulations.

Note that independent of the model versions described in the previous section, we always compute MSE based on all state variables $h$, $Q$ and $Q_w$ for all nodes and links. In addition, we distinguish:

- training series, based on which MSE values are computed as loss-function for parameter estimation;
- validation series, i.e. an independent rainfall series for which MSE is computed in each iteration of the parameter optimization, but which is solely used for validating that no overfitting takes place; and
- testing series, i.e. a third, independent rainfall series which is never presented during training, and which is used for evaluating model performance.

To ensure computationally efficient training in an operational setting, we divide the training series into so-called windows. The surrogate model is initialized from the HiFi simulation results at the beginning of a window and then independently simulates the hydraulic states for a number of time steps corresponding to the so-called window size. In this way, the training series is divided into chunks that can be processed in parallel (independent from each other). In the extreme case, a window size of one time step maximizes the potential for parallel processing and will thus yield the fastest training. However, it also entails a high risk of overfitting, because the surrogate is initialized from the true values at each time step, and thus never forced to generate stable simulations in the training phase.

## 2.4 Technical Implementation

Surrogate models were implemented in Python 3.8 using Google's Tensorflow library version 2.7.0 (Abadi et al., 2016). Parameter estimation was performed over 2,000 epochs (or optimizer iterations) using the widely applied Adam algorithm (Kingma and Ba, 2014). The algorithm was (by trial and error) configured with an initial learning rate of 1e-3 which exponentially decayed to 1e-4 (You et al., 2019). We stopped training preliminarily if the validation loss did not decrease for 500 epochs in a row.

## 2.5 Design of experiments

### 2.5.1 Catchment and HiFi model

We tested our proposed emulator on a medium-sized test system extracted from a real UDS in Bochum, Germany (Figure 2). This multi-branch system counts a total of 60 pipes and 61 nodes collecting combined sewage from several catchments, covering a total area of approximately 140 ha. Except for the outlet, all nodes in the model are equipped with a spilling weir placed near ground level to simulate water discharged to the surface. For node 27793 the weir crest was placed just above the top of the pipe level to simulate frequently occurring overflows. The last node is an unconstrained outlet. Except for the outlet, each node is connected to a single catchment, each having a different area and imperviousness. Detailed system properties are provided in the supporting information.

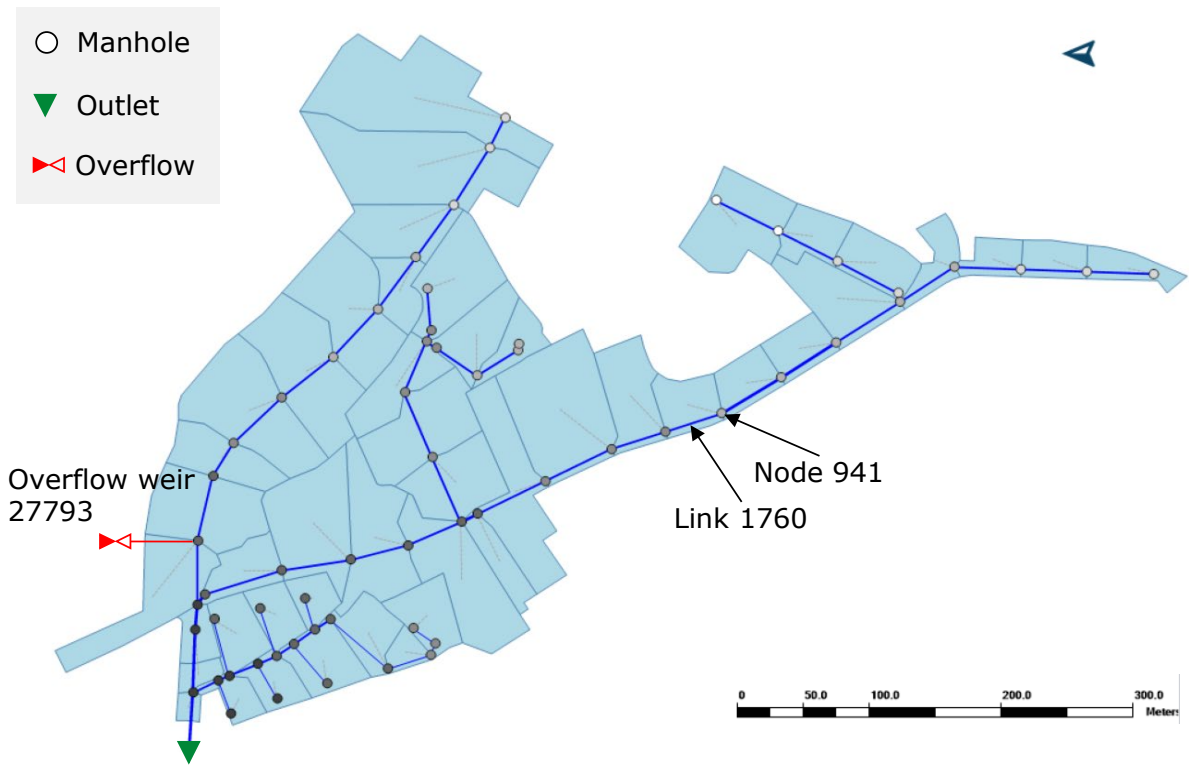

Figure 2. Map view of test system, with nodes (grey dots), pipes (dark blue lines) and catchments (blue polygons). Locations of the combined sewer overflow and downstream outlet are shown on the map. Nodes with lower invert elevation are marked with darker color.

### 2.5.2 Training, validation and testing data

We generated training, validation and testing data for the development of surrogates by performing hydrodynamic simulations with the MIKE 1D engine (DHI, 2021). For surrogate training and validation, we performed hydrodynamic simulations using two different sets of rainfall series as input:

A. In dataset A, we extracted rain events from a continuous 40 year series of rainfall observations in 1 minute resolution, assuming that a rain event ends when the observed rain intensities did not exceed 0.02 mm/min for at least 120 minutes. The rain gauge is part of the Danish SVK network and located in Odense, Denmark. From the identified events, we manually selected 154 rain events for training and 43 rain events for validation. In both cases, we aimed to include events with different temporal variations and intensities. The selected events were concatenated into one time series, resulting in a rainfall series for training that consisted of 73,990 one-minute data points, and a validation series of 19,095 data points.

B. In dataset B, we aimed to create a rainfall series where extreme events were overrepresented to ensure that the surrogates can appropriately learn the dynamics of these phenomena. We repeated the above process, considering 10 rain gauges from various locations in Denmark, each with 40 years of rainfall observations. Once events were identified for each gauge, we manually selected 203 events for training and 58 events for validation from the pool of events, focusing our selection on intense rainstorms. In particular, we included all extreme rainstorms where flooding would be likely to occur. The selected events were again concatenated into one series, resulting in a training series of 74,769 time steps and a validation series of 21,700 time steps.

All surrogates were tested by performing simulations for a rainfall series consisting of 40 years of continuous observations in Næstved, Denmark. This series was not considered in the creation of datasets A or B. The gauge is located approximately 30 km away from the nearest gauge considered in the creation of dataset B and was intended as a fully independent test series. The testing series included approximately 8,200 rain events.

An overview of the rain events included in the different datasets is included in the Supporting Information.

### 2.5.3 Summary of Experiments

The technical implementation of our surrogates has two main hyperparameters. These are the window size used for subdividing the training data series, as well as the complexity of the neural network $N$ that is used to model the residue of the linear predictor $L$ (see Eq. 5). To identify a reasonable configuration, we compared $MSE$ obtained on the validation dataset (so-called validation loss) and training times for the hyperparameter configurations shown in Table 1. This process was repeated five times for each combination of hyper-parameters, because different (random) initializations of the parameters in the neural networks lead to slightly different validation losses.

Subsequently, we selected the configuration with the best compromise between validation loss and computational efficiency, and compared surrogate versions with and without physical constraints that were trained on datasets A or B.

Table 1. Design of experiments. A reasonable complexity of the neural network and an appropriate training window size were selected in the first stage. Subsequently, performance of different surrogate versions was assessed.

|  | Physical constraints | Training series | Window sizes [min] | Neural network $N$ specifications |
|---|---|---|---|---|
| **Stage 1** Selection of hyperparameters | Without, With | A, B | 1, 10, 60, 120, 360 | S1 = 2 hidden layers of 10 neurons each<br>S2 = 4 hidden layers of 20 neurons each<br>S3 = 6 hidden layers of 50 neurons each<br>S4 = 6 hidden layers of 100 neurons each |
| **Stage 2** Assess performance and effect of physical constraints and training series | Without, With | A, B | 60 | S4 |

## 2.6 Scoring criteria

To compare the surrogate simulations against HiFi simulations, we considered widely used scoring criteria that were computed for the testing series. We considered RMSE and $R^2$ to measure both average and event-based performance of the surrogates. All scores were computed in the untransformed space, i.e. the units of RMSE are meter for water levels and $m^3/s$ for flows.

We evaluated the total excess flow volume $\sum_t Q_{w,i,t}$ for individual nodes during individual rain events, to assess whether overflows and surcharges were predicted in the correct locations and with accurate magnitude

Finally, to evaluate whether the model results are biased, we compared total runoff, surcharge, overflow and outflow volumes.

# 3 Results

## 3.1 Hyperparameter selection

Figure 3 illustrates the validation loss (smallest $MSE$ obtained on the validation series during training) for different combinations of neural network configuration and window size, considering surrogates that included physical constraints for surcharges and that were trained on series A. The results for this surrogate configuration illustrate the considerations arising during training. Results for other surrogate configurations are included in the supporting information.

Smaller window sizes allow faster training of the surrogates when parallel processing is exploited but increase the risk of overfitting. This issue is very clear from the figure, where models trained with window sizes of 1, and in some cases 10 minutes yield substantially larger validation losses. Surrogates with too few parameters lead to unstable validation results because they have difficulties learning the dynamics of the system. Considering also the results obtained for data series B (supporting information), network sizes S3 and S4 are preferable. Moving from neural network size S1 to S4, computation times per epoch on a single CPU increased moderately (in the order of 30%). Considering the more stable convergence properties of the more complex models, we moved forward with network size S4 and a window size of 60 minutes for the remaining study.

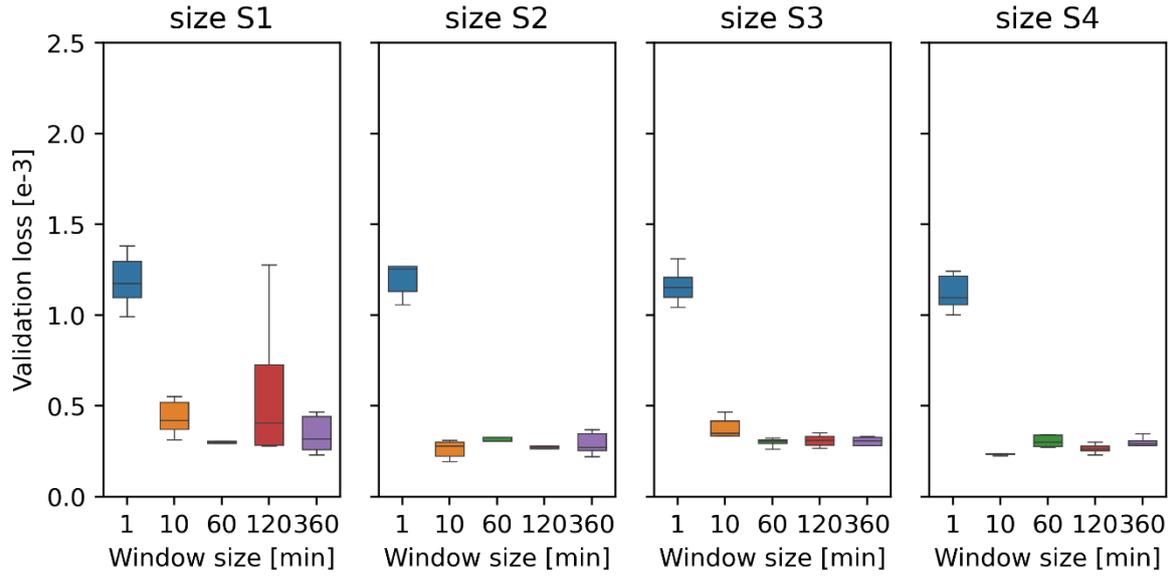

Figure 3. Box plots of validation loss as a function of network size and input window size. The model included physical constraints on surcharges and was trained with rain events A (each configuration was tested 5 times).

## 3.2 Hydrographs

Figure 4 illustrates time series for link flows, water levels, overflows and surcharge flows in the locations indicated in Figure 2. Time series are shown for a single, intense rain event in the testing series. Time series plots for all links and nodes are included in the Supporting Information. The presented surrogate results were generated from surrogates trained on data series B. Amongst the 5 training iterations for each model (Figure 3), we show results for the best performing model without physical constraints (lowest validation loss) and the worst performing model with physical constraints (highest validation loss).

Comparing the states simulated by the HiFi model, as well as the surrogates with and without physical constraints, we can observe some general trends: 1) Flows and water levels are simulated with high accuracy by both surrogate models. 2) The surrogate without physical constraints creates false predictions of excess flows $Q_w$ with both positive and negative signs. These issues are more pronounced for surcharges (panel D) than for overflows (panel C). The latter occur more frequently in the dataset and are thus easier to learn for the surrogate. The issue is avoided when introducing physical constraints for surcharge predictions into the surrogate.

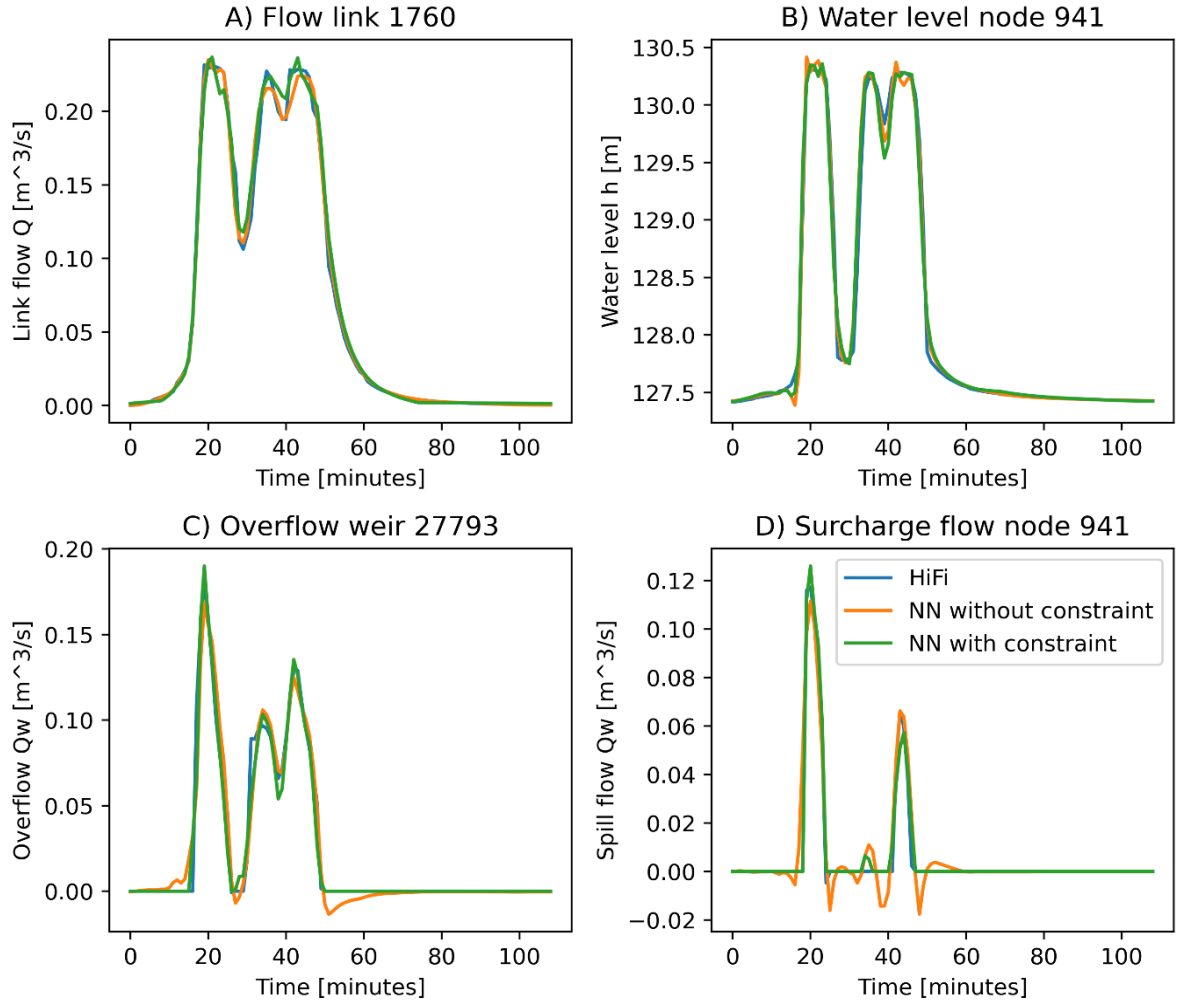

Figure 4. Time series plots for a rain event in the test dataset, comparing HiFi model simulations (blue), against neural network (NN) emulators without (orange) and with (green) physical constraints. The surrogates were trained using rain series B and network size S4.

### 3.3 Event-based accuracy

Figure 5 shows RMSE computed for $h$, $Q$, $Q_{w,overflow}$ and $Q_{w,surcharge}$ for individual rain events (considering the same surrogates as in Section 3.2). We used the peak flow simulated by the HiFi model at the outlet to distinguish low and high flow conditions in the figure. A general increase of RMSE can be observed for high flow conditions. This may be linked to a change of dynamics in extreme flow situations and the relatively fewer occurrences of such situations even in data series B. Nevertheless, RMSE values for levels and flows are in the range of few cm and few l/s, respectively. This suggests that the surrogates generally achieve high accuracy. RMSE values for water levels and flows are slightly higher for the model considering physical constraints. In this model configuration, the simulated water levels are tied to the activation of mass balance computations for excess flows. Thus, the surrogate has less freedom to fit the water level data than the model without constraints. In contrast, the model including physical constraints achieves much lower RMSE values for $Q_{w,overflow}$ and $Q_{w,surcharge}$ (panels C and D). Deviations in low flow conditions (where the HiFi model does not simulate excess flow) are entirely removed. In addition, smaller RMSE values are also achieved in periods where excess flows occur.

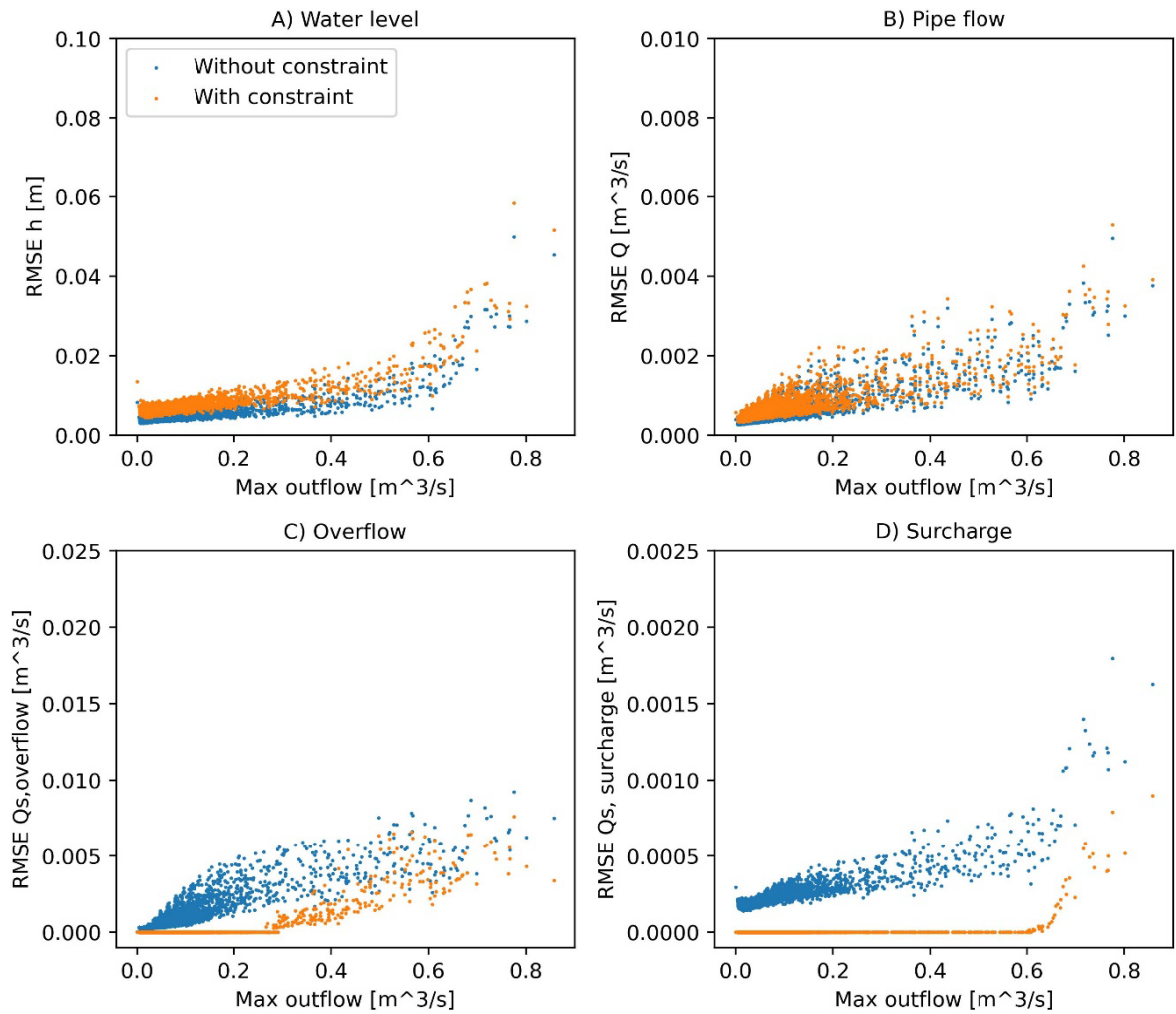

Figure 5. RMSE of water level in nodes (A), discharge in pipes (B), discharge over overflow weir (C) and surcharge (D) as a function of event intensity, expressed as maximum discharge at the outlet. Each dot represents a different rainfall event in the testing dataset. The results are shown for the emulator trained on data series B, with and without physical constraints.

Figure 6 analyses the behavior of predicted overflows and surcharges more in depth. The figure compares total overflow and surcharge volumes for individual rain events in the testing dataset, considering each node individually. Panels A and C consider surrogates trained on data series A, while panels B and C consider surrogates trained on data series B. Different colors distinguish surrogates with and without physical constraints.

Comparing the two surrogate configurations, we can conclude that the introduction of physical constraints leads to a much better representation of excess flows. However, comparing panels A and C against panels B and D, a clear effect of the training series is also noticeable. When trained on series A, predictions of overflow and surcharge volumes are slightly more uncertain, because these events are too sparse in the training data. The predicted water levels and link flows are then not properly tuned to enable an accurate mass balance calculation. Considering training series B, a linear relation is achieved between overflow and surcharge volumes predicted by the surrogate and the HiFi model.

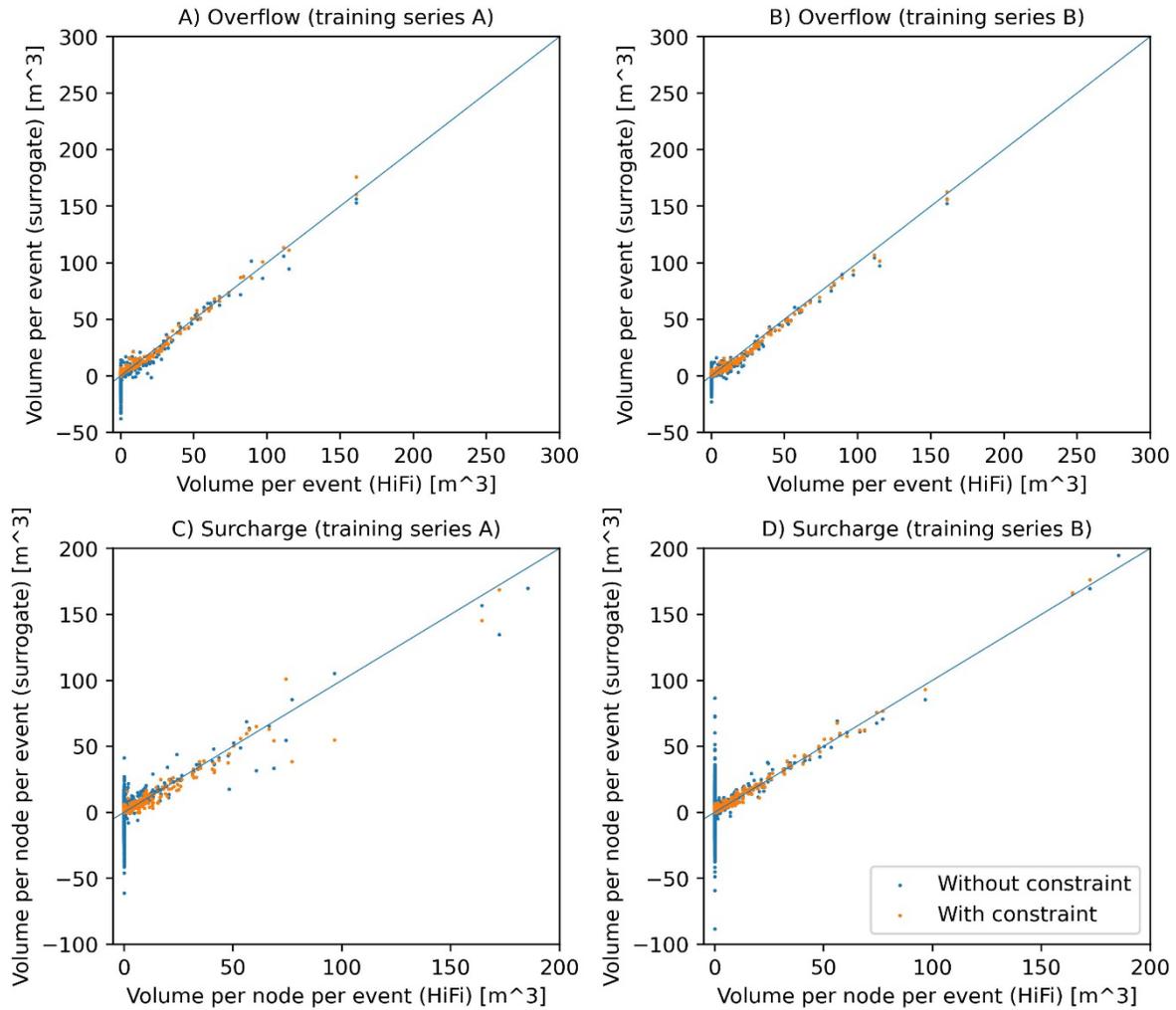

Figure 6. Panels A and B: Total overflow volume simulated for weir 27793 in individual rain events by the HiFi model as well as by the emulator trained on datasets A and B. Panels C and D: Total surcharge volumes simulated by the HiFi model, and the emulator trained on datasets A and B. Each dot represents the surcharge volume simulated for an individual node for and individual rain event in the testing dataset.

### 3.4 Time-averaged test scores

The accuracy of the emulator predictions was assessed by comparison with the hi-fi model simulation over the testing events. The assessment was conducted separately for each type of hydraulic state (Figure 7). The RMSE of the simulated water level and discharge in pipes was similar across the four configurations tested. As already indicated by the previous results, imposing physical constraints led to improved simulations of surcharges and overflows, which is indicated by improved RMSE values (panel A) and much improved $R^2$ values (panel B) and reduced volume errors (panel C). Comparing the two model configurations with physical constraints, the performance of the surrogates trained on different data series was very similar. However, surrogates trained on series B yielded slightly lower RMSE values and slightly higher $R^2$ values.

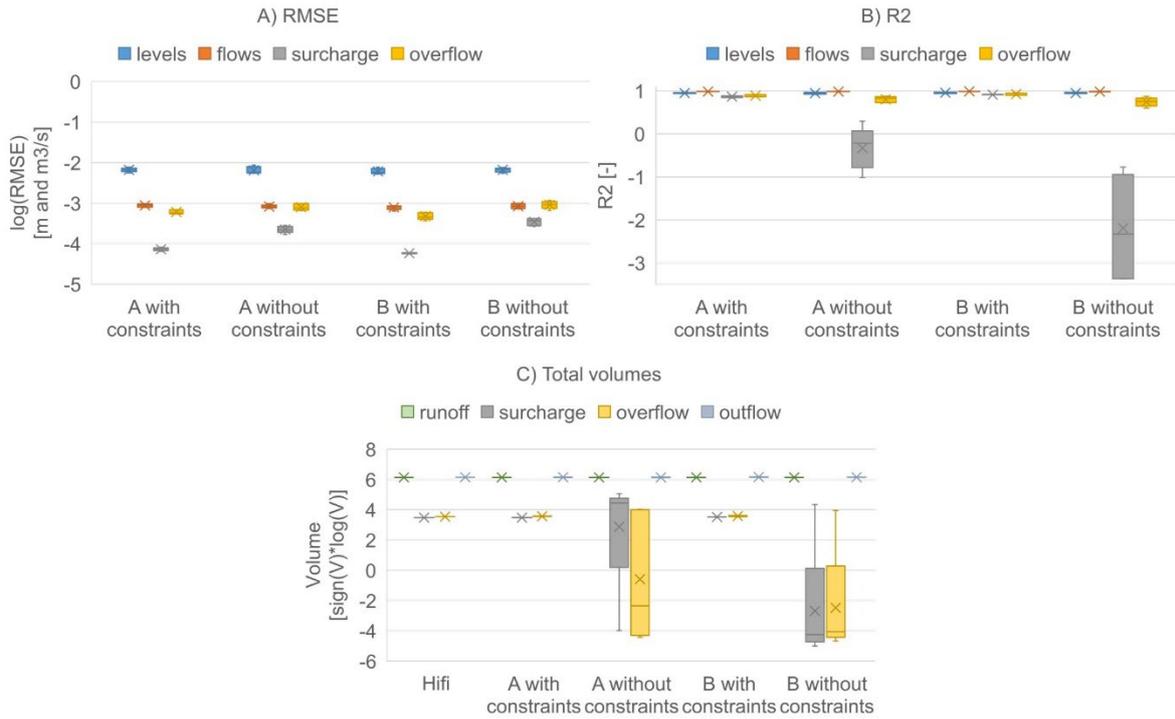

Figure 7. Evaluation scores of emulator predictions for the test dataset, divided by metric (A: Root Mean Square Error, B: $R^2$) and hydraulic state (water level in nodes, flows in pipes, surcharge and overflow) for all four tested configurations. Panel C shows the total runoff, surcharge, overflow and outflow volumes simulated for the testing series. Panel C shows log-transformed absolute values that were multiplied by the sign of the original value.

Figure 8 illustrates for the surrogate with physical constraints how the accuracy of simulated water levels and link flows varies in space. While the surrogate generally achieves very high $R^2$ values, model performance is bad in some upstream links and nodes in the center of the catchment. In this area, the connected sub-catchments have very low imperviousness ratios. In the provided system model, the runoff computation for pervious areas considered initial losses of 10 mm and Horton infiltration capacities that varied between 72 and 18 mm/hr. This implies that water rarely flows into these parts of the sewer system and the surrogate cannot properly learn the relationship between inflow and hydraulics. Training a surrogate for pipe hydraulics is not limited to considering the rainfall-runoff characteristics of the existing system. The issue can therefore be avoided by ensuring that training data are generated in such a way that high and low inflow situations occur in all parts of the system.

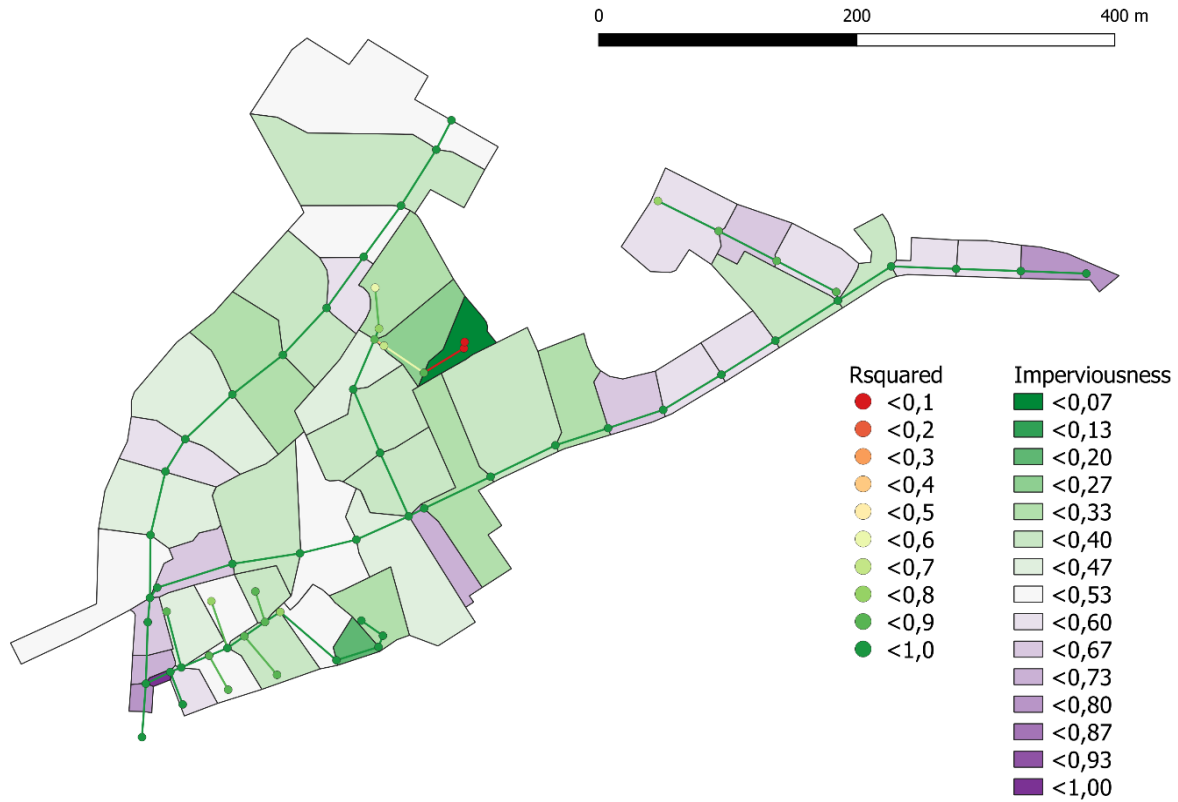

Figure 8. Average R$^2$ values for node water levels and link flows considering the surrogate with physical constraints trained on series B. Catchments are colored according to which portion of the catchment area is impervious.

## 3.5 Computation time

Table 3 compares computational effort for training and running the surrogates against the time required to simulate the testing series in the HiFi model. The surrogates are currently trained against results from a HiFi simulations. Therefore, we also included the time required to perform these simulations for the training and validation dataset ("label generation"). Tests were performed on an HP Zbook G9 with Intel Core i9-10885H processor and 32GB RAM. Both HiFi simulations and surrogate training and simulation were performed using a single CPU only.

Training times ranged between 0.5 and 1.2 hours. Surrogates trained on data series A generally converged faster and were therefore not trained for the full 2,000 epochs. This effect may be related to the less complicated dynamics in this data series. Surrogates including physical constraints were in the order of 10% faster than surrogates without physical constraints, because the surrogate simulation was performed without $Q_w$ and thus included ~33% fewer state variables. Excess flows were instead computed efficiently in a postprocessing step.

Prediction times for the testing series (8,200 rain events) were in the order of 80 s and thus in a range that enables real time applications as well as interactive simulations in workshop settings. For comparison, HiFi simulations for the testing series lasted 6,200 seconds, considering a fixed simulation time step of 5 s. Note that the simulation time step of a HiFi model is system dependent. Short pipes with lengths in the order of 10 m or complicated hydraulic structures can require even shorter simulation time steps and thus increase simulation time. On the other hand, HiFi simulation times can, in the experience of the authors, be in the order of 2 to 5 times less than shown in Table 3, if the pipe system does not involve such complicated hydraulic situations, and if adaptive time stepping schemes (DHI, 2021) are employed. Surrogate simulation times will remain constant as long as the same simulation time step (here one minute) is considered. However, prediction accuracy may be reduced in some situations.

While we tested computation times in a single CPU setting, the surrogate models can exploit parallel computing, which is important for implementation. In our tests, we did not achieve reduced simulation times by employing GPUs. This is most likely due to the autoregressive character of the models, which implies that the technical implementation includes a loop over a time series. The higher processing speed of CPUs outperforms the parallelization capacities of GPUs in this situation.

Table 2. Computation time for training surrogate models (includes label generation and actual training process) and simulating the testing series (8,200 rain events; 1.5E6 time steps). Surrogate training and simulation times are averaged across 5 model runs. HiFi simulations considered a fixed time step of 5 seconds. All time estimates were determined using a single CPU.

| Rain | Model | Label generation [s] | Training [s] | Simulation time [s] |
|---|---|---|---|---|
|  | Hi-fi |  |  | 6,200 |
| A | Emulator (wo phy.) | 310 | 2,700 | 80 |
|  | Emulator (w phy.) | 310 | 2,020 | 73 |
| B | Emulator (wo phy.) | 310 | 4,434 | 80 |
|  | Emulator (w phy.) | 310 | 3,940 | 74 |

# 4 Discussion

## 4.1 Results and limitations

The results shown in the previous section illustrate that our surrogate modelling approach can simulate water levels, flows, overflows and surcharges with high accuracy. The surrogates enable long time series simulations (in our case 8,200 rain events extracted from 40 years of rainfall observations) of water levels, flows and surcharges in all nodes and pipes of a drainage system within few minutes. While these results are promising, our work has several clear limitations that will need to be addressed in further research to make the approach applicable in practice:

1. While our test case is an example of a typical drainage network, it does not include any structures apart from overflow and surcharge weirs, and all pipes were circular. Very irregular cross sections (e.g. when simulating open water bodies) may create dynamics that are challenging to learn and that lead to lower surrogate performance. Similarly, sewer systems often include trunk sewers that can be dominated by inflows in few selected points, rather than catchment inflows that are somewhat uniformly distributed across the nodes of the drainage system. These situations may require modified surrogate architectures and training configurations.
2. Applying our surrogate approach in practice will very likely require a subdivision of large drainage networks into many small subsections, each of which is simulated by an independent sub-model (see Section 4.2). Implementing such an approach requires that sub-models can be trained independent from each other, and subsequently generate accurate simulations for the combined systems. Our surrogate architecture certainly enables interactions between sub-models. For example, water level predictions of one model can be imposed as a boundary condition for another model to enable the simulation of backwater effects. However, in this paper we have neither documented approaches for training sub-models individually, nor assessed the accuracy of simulations in this setting.

3. We created artificial rainfall series for training the surrogates. These were defined based on hydrological intuition with the aim of representing the range of relevant dynamics. We have not investigated how many data points should be included in the training series. Structured approaches for selecting the rain events included in the training data (e.g (Allen et al., 2022)) may enable surrogate training with shorter data series and therefore also shorter training times.

4. Figure 4 suggests a clear dependency between the complexity of the neural network included in the surrogate model and the accuracy of the simulations. In general, we would expect that an increase of the number of state variables in the model requires that more complex neural networks with more parameters are included in the surrogate to achieve sufficient accuracy. The form of this relationship is currently unknown.

5. All simulations in our study were performed considering a uniform distribution of rainfall in space. Considering the size of the test case, this assumption is quite realistic. However, surrogate developments for larger drainage systems will need to consider spatial rainfall variations.

6. While pipe flow simulations are performed with high accuracy, we are not currently enforcing mass balance in the model architecture.

## 4.2 Upscaling to large drainage networks

We envision that our surrogate approach will have two main applications:

1. Early design phases of urban drainage systems, where multiple stakeholders in, e.g., workshop settings want to obtain qualified bids for how urban hydrology is affected by urban development and water management measures, and
2. Real time applications, where drainage system operators need fast models to monitor and control drainage system operations.

Surrogate training times are critical for design applications because they increase the waiting time until updated results become available, and thus limit the ability to use the models interactively. In its current configuration, our surrogate approach enables assessments of how changes in urban runoff affect sewer overflows, surcharge frequencies and wastewater treatment plant inflow within a few minutes and thus sufficiently fast. However, any modifications of the hydraulic behavior of the pipe system will require retraining of the surrogate models, leading to waiting times in the order of one hour. However, stakeholders will often want to investigate the effect of changing few selected pipes in a well-confined area. For this reason, we envision that our surrogate approach should be used in a setting where the drainage system is divided into many small subsystems, each of which is represented by an independent surrogate. This approach is quite similar to the structure of existing conceptual hydrological models that are used in urban hydrology (Kroll et al., 2017; Thrysøe et al., 2019). It implies that only one or few surrogates for small subsystems need to be retrained if changes are made to a subsystem, and not a surrogate of the entire drainage network that often involves thousands of pipes. It is currently unclear what should be the optimal size of these subsystems.

Training times are less relevant for real-time applications because the layout of the drainage system typically does not change. However, subdividing the system may also be beneficial for these applications, because it enables executing simulations individually for each sub-model on a time step to time step basis. Compared to a very large surrogate with thousands of states, this reduces memory requirements, and may also speed up simulations if sub-models are processed in parallel.

## 4.3  Research perspectives

We anticipate that future research can much improve our surrogate approach by reducing training times and incorporating additional physical constraints into the model architecture. In terms of reduced training efforts, graph neural networks (Zhou et al., 2020) enable the creation of surrogates where new states for a node or pipe are predicted based on the current state of its neighbors. While the approach presented in our paper considers all states in the drainage network as input to a single neural network, graph approaches would process each node and link individually. When combined with physical system properties such as pipe lengths and slopes, the resulting surrogates may become transferable across catchments, reducing training efforts to near zero. Another option for streamlining the training process is to use physics-informed loss functions during surrogate training (Raissi et al., 2019; Wang et al., 2021). This approach would allow us to train surrogate models without generating training data in a hydrodynamic model first. While physics-informed loss functions increase training times, it may be attractive from a practical viewpoint to avoid data management and stability issues associated with automated handling of a numerical engine. In addition, physics-informed loss functions may enable bypassing some of the numerical simplifications that are commonly implemented in commercial software packages for drainage system simulation. Finally, transfer learning (Jin et al., 2021), i.e., initializing the surrogate parameters not randomly but based on experience from previous training iterations is likely to enable surrogate training in much fewer epochs than documented in our paper.

With our physics-constrained surrogate configuration we have far from exploited all opportunities for implementing physical knowledge into the surrogates. The architecture is flexible in terms of considering other surrogate formulations. For example, flow formulas were used in the development of cellular automate for drainage networks (Austin et al., 2014) and could be incorporated into our the $L$ term in Eq. 3 to generate robust flow predictions. Another relevant consideration would be to compute volume changes in each node and link during each simulation time step, and to penalize the surrogates for mass balance deviations during training.

Finally, the model configuration presented in Eq. 3 is generic and can be applied to other water systems. (Wandel et al., 2021) used a similar approach to simulate 3D fluid flows for graphics applications, suggesting that we can derive similar surrogates for, for example, flows in secondary clarifiers in wastewater treatment or 2D surface flows in flood situations.

# 5 Conclusions

We presented a new surrogate approach for simulating pipe hydraulics, based on physics-guided machine learning using generalized residue networks. We anticipate that this approach will complement existing numerical models in initial design phases for drainage systems and real-time applications, where fast simulations are required. Based on our results, we draw the following conclusions:

1. We can create surrogate models that accurately simulate water levels, flows and surcharges in all nodes and links of a drainage network based on generalized residue networks.
2. Compared to a state-of-the-art numerical simulation engine, simulation times for long rainfall time series of several years are reduced between one and two orders of magnitude. This enables fast and detailed drainage simulations in interactive workshop settings and in real time applications.
3. Surrogate training times for a drainage system with ~60 links are currently in the order of one hour, considering a single CPU. Transfer learning approaches and graph neural networks will likely enable significant reductions of training efforts in practice.
4. Implementing physical constraints in the surrogate in the form of mass balance calculations for surcharges improves performance compared to a purely data-driven approach.
5. Surrogate training needs to cover the entire range of hydraulic situations for which the surrogate is intended to be used. Extreme events need to be oversampled in the training data to ensure that surrogates can accurately learn the dynamics of these situations, and a sufficient variation between high and low inflows needs to be applied to all parts of the system.

Generalized residue networks are a generic technique for time-dependent modelling. Their application for dynamic simulations of other types of water systems is therefore an interesting research avenue that warrants investigation.

# 6 Acknowledgements

This work was funded by the Danish Environmental Agency (Miljøstyrelsen) under the MUDP programme through the Clacos project (grant number 2020-15748). We thank Ralf Engels, Michaela Ringelkamp and Marko Siekmann from the City of Bochum for providing system data as well as sparring during model development. We thank Peter Steen Mikkelsen for proofreading the manuscript and for his support in initiating the project.